\documentclass[11pt]{article}

\usepackage[final]{acl}

\usepackage[T1]{fontenc}
\usepackage[utf8]{inputenc}
\usepackage{times}
\usepackage{latexsym}

\usepackage{microtype}
\usepackage{inconsolata}
\usepackage{graphicx}
\usepackage{subcaption}
\usepackage{booktabs} 
\usepackage{textcomp} 
\usepackage{tikz} 
\usetikzlibrary{arrows.meta, positioning, fit}

\usepackage{hyperref}

\definecolor{darkblue}{rgb}{0, 0, 0.5}
\hypersetup{colorlinks=true, citecolor=darkblue, linkcolor=darkblue, urlcolor=darkblue}

\usepackage{amsmath}
\usepackage{amssymb}
\usepackage{mathtools}
\usepackage{amsthm}

\usepackage{url}
\usepackage{multirow}
\usepackage{adjustbox}
\usepackage{caption}
\usepackage{todonotes}
\usepackage{listings}
\usepackage{enumitem}
\usepackage{seqsplit}
\usepackage{wrapfig}
\usepackage{tcolorbox}
\newtcolorbox{prompt}[1][]{
  colback=gray!10,        
  colframe=gray!50!black, 
  nobeforeafter,
  title=Prompt,
  #1 
}
\usepackage[flushleft]{threeparttable}

\usepackage[capitalize,noabbrev]{cleveref}

\usepackage[normalem]{ulem}
\newcommand{\modify}[1]{\textcolor{black}{#1}}

\newcommand{\added}[1]{\textcolor{black}{#1}}

\newcommand{\edit}[1]{\textcolor{black}{#1}}

\newcommand{\rev}[1]{\textcolor{black}{#1}}

\title{\added{Scalable Supervision for Software Agents via Patch Reasoning}}

\author{%
    \textbf{Junjielong Xu}$^1$\thanks{Work was done when interning at ByteDance TRAE.}
    \quad \textbf{Boyin Tan}$^1$
    \quad \textbf{Xiaoyuan Liu}$^1$
    \quad \textbf{Chao Peng}$^3$
    \quad \textbf{Pengfei Gao}$^3$
    \quad \textbf{Pinjia He}$^{1,2}$\thanks{Correspondence to: Pinjia He <hepinjia@cuhk.edu.cn>.}
    \vspace{1em}
    \\
    $^1$School of Data Science, The Chinese University of Hong Kong, Shenzhen\\
    $^2$School of Artificial Intelligence, The Chinese University of Hong Kong, Shenzhen\\
    $^3$ByteDance 
}

\begin{document}

\maketitle

\begin{abstract}
While language model agents have advanced software engineering, existing test-based supervision is limiting its scalability on real-world issues.
The reason is twofold: (1) high-coverage tests are naturally rare in the wild,
and
(2) building and running test sandbox is heavy and fragile.
To unlock supervision scaling, we
propose R4P, a reasoning-based method that provides scaffold-agnostic rewards.
{R4P uses a group-wise training objective, enabling it to verify multiple patches against each other's modification and gain a dense reward for supervising agents without executing tests or relying on specific agent trajectories.}
R4P achieves 72.2\% Acc.\ for verifying patches from SWE-bench, {competitive with proprietary models}.
To show the downstream practical utility of R4P,
we design and train {an execution-free} scaffold, Mini-SE, with \textit{pure} RL via R4P.
Mini-SE achieves 26.2\% Pass@1, showing a 10.0\% improvement over the original Qwen3-32B, and can be further improved to 32.8\% with R4P for test-time scaling on patch selection.
The stable scaling curves illustrate that though imperfect, R4P can still reliably support downstream tasks at scale.

\end{abstract}

\section{Introduction}\label{sec:intro}

Large language models (LLMs) agents have made notable progress in software engineering (SWE) thanks to rule-based reinforcement learning (RL) and test-time scaling (TTS)~\citep{jimenez2023swe,deepswe2025}.
Since the output of SWE (patches) are always \textit{non-unique} and hard to \textit{formally verify}~\citep{ter2024formal}, \textit{testing} was adopted as a proxy reward.
However, while a small number of carefully maintained codebases have relatively well-tested issues and are widely-used for existing datasets~\citep{jain2025r2e}, the vast majority of the tests in the open-source community suffer from \textit{low-coverage problem}~\citep{yu2025utboost}. We even found that only 28.11\% of resolved issues have corresponding tests on projects with over 500 stars (Appendix~\ref{app:pre}). Moreover, building test sandboxes (e.g., Docker) for each instance is highly labor-intensive~\citep{pan2024training,jain2025r2e}, and hosting them for parallel rewarding is unstable due to surge workload~\citep{deepswe2025}.
Thus, we consider that the bottleneck is not data availability but \textit{supervision scalability}, and model-based verifiers \modify{that judge without executing tests} are needed.

Ideally, a general reward model~\citep{skyworkcritic2024,chen2025rm} could serve as a scalable verifier.
However, they are trained to judge the \textit{relative} preference of solutions than their \textit{absolute} correctness.
Since the labels are binary (Pass/Fail) but the patches can be non-unique and diverse, it is hard to model an order between two different correct patches.
Furthermore, without test execution, LLMs lack the dependency context to catch subtle bugs in isolation.
Existing SWE verifiers~\citep{pan2024training,jain2025r2e,deepswe2025} are trained to use certain agent's trajectories as an additional context.
However, it creates a strong coupling between the model and a specific interaction style, hindering its ability to generalize to other agents.

We introduce \textit{R4P}, a
reasoning-based, scaffold-agnostic patch verification method for \textit{any} SWE agents: \modify{when rewarding,} it does not rely on any golden test, developer patch, agent trajectory, or run-time sandbox.
We consider that patch verification is fundamentally a reasoning task which can be improved via RL. This mirrors how real-world maintainers review patches based on their knowledge and a detailed analysis of the patch, rather than writing and running new test scripts~\citep{baum2016faceted}.
R4P adopts a group-wise training objective:
for a given issue and a set of candidate patches, R4P is trained to assess each patch by comparing it against others in the group for mutual context information, compensating for the absence of tests and facilitating the detection of subtle errors via using other patch content.
This avoids introducing redundant context tokens (e.g., agent trajectory) and enables verifying multiple patches within one forward computation.
Furthermore, R4P uses the accuracy of each group as the reward, providing a denser signal ([0,1]) than naive binary classification (\{0,1\}).
This also reduces the threat of rewarding incorrect reasoning chains that happen to arrive at the correct answer, as coincidentally correct labels across a group are far less probable than in a single instance. This leads to a more stable training process with improved convergence.
We evaluate R4P on patches generated by four different agents from SWE-bench-verified Experiments. When built upon a 32B model, R4P achieves 72.2\% Acc., \modify{comparable to proprietary models like Claude-4.5-Sonnet}.

{To further show that R4P is practical for downstream tasks,}
we developed \textit{Mini-SE}, a light-weighted, execution-free agentic scaffold that excludes any impact from test execution.
We use R4P to train Mini-SE based on Qwen3-32B and R2E-Gym issues without using official test sandbox.
It shows a stable scaling curve on both reward and test score along with the data scaling, illustrating that R4P's imperfect yet scalable supervision is practical for agentic RL without reward collapse.
This confirms our core hypothesis: \textit{the benefits from data scaling of R4P can outweigh the losses from slightly noisy signals}.
While Mini-SE does not test its generated patches during rollout, its verification burden is transferred to R4P for test-time scaling:
when using R4P for patch selection, the final score can be improved to 32.8\%.
In addition, R4P verifies each patch within a second on average, much faster than the minute-level time cost of testing.
In conclusion, \textit{while R4P is not a replacement of high-quality tests \modify{(its training itself relies on high-quality tested patches)}, we believe R4P is a practical and scalable alternative for SWE agents when well-maintained data is unavailable}.

\section{Background}\label{sec:pre}

\textbf{LLM for Issue Resolving}: Software Engineering (SWE) tasks have increasingly gained attention, especially for real-world repository-level issue resolving tasks~\citep{jimenez2023swe}. Given an issue description in natural language (e.g., a feature request or a bug report), the models need to scan the whole repository to understand the problem, locate the place for editing, and finally submit a patch as the answer. The golden tests (i.e., the unit tests submitted in the developer patches) and corresponding sandbox environments (typically based on Docker) will then be applied to verify the patch.
While such test-based verification is effective for benchmarking, it cannot provide scalable supervision for improving model's SWE capability.
We identified two major problems:

\textit{P.1: Test coverage problem}:
In practice, unit tests are unavoidable to fail to cover all edge cases, allowing tricky patches to pass without genuinely solving the issue~\citep{mockus2009test}.
Though a small volume of data from actively-maintained codebases includes relatively reliable tests and has been used in existing datasets~\citep{jain2025r2e}, it is challenging to replicate this reliability when scaling up to the wider open-source projects.
Furthermore, our study shows that 28.11\% issues from projects with more than 500 stars even do not contain \textit{any} tests, let alone the test coverage problems (see Appendix~\ref{app:pre}).

\textit{P.2: Testing is heavy}: Even with the help of LLM, building a single issue's Docker image takes about 7 human minutes~\citep{jain2025r2e}. Thus, existing datasets contain less than 5,000 real-world executable issue instances~\citep{pan2024training}. While \textit{issue synthesis} allows for data scaling in a reusable sandbox via commit backtranslation~\citep{yang2025swe}, they are proven to be much less effective than real-world issues~\citep{deepswe2025}.
Furthermore, maintaining hundreds of Docker containers (e.g., 64 batch size $\times$ 8 rollouts) for parallel online rewarding is highly unstable, as containers are prone to crashing under peak load~\citep{deepswe2025}.
\textbf{Challenges of Model-based Verification}:
A way to scale supervision is to train a reward model (RM) as patch verifier, which judges LLM responses without relying on tests and environments. However, it is usually infeasible in practice to incorporate RMs into SWE due to two challenges:


\textit{C.1 Lack of sufficient reference}: In addition to scalar and trajectory RMs discussed in Sec.~\ref{sec:intro}, we explored the capability of advanced LLMs as RMs for patch verification (see Sec.~\ref{sec:analysis}).
Our analysis of failure cases reveals the reason is due to the lack of dependency context for the patch, preventing the model from identifying subtle incompatibilities or bugs with the existing codebase.
While agentic search is a potential solution, precisely identifying all relevant context is not practical, as an issue in a large project can be far-reaching. For most of the cases, it is as costly as patch generation but still inaccurate~\citep{meng2024empirical}.


\textit{C.2 Sparsity of outcome space}: Since the output space of patch verification is binary (Pass/Fail), the supervision signal for RM training is very sparse, making naively mapping input issue-patch pairs to a Pass/Fail label very challenging. In practice, diverse patches can solve the same issue without a dense, relative relationship between each other, which further precludes the use of dense training objectives like Bradley-Terry model~\citep{kendall1940}. 
Furthermore, binary rewards are easily obtained by chance without sound reasoning, making the training unstable for RL.

\section{R4P}\label{sec:r4p}


To provide scalable patch verification for SWE tasks, we propose R4P, a reasoning {patch verification method} trained via pure rule-based RL.
We consider that patch verification is naturally a reasoning task, as human developers review pull requests (PRs) by reasoning about the static code changes, rather than designing new, issue-specific golden tests for every submission.
{R4P enables efficient data scaling for unlabeled or low-test-coverage real-world issues, thereby supporting continuous learning beyond the limited well-tested data.}

\textbf{Task Formulation} To address the challenges of reference deficiency (\textit{C.1}) and sparse rewards (\textit{C.2}), R4P introduces a group-wise task formulation.
Given a SWE issue $I$ in issue description and a group of patches $P=[p_1,p_2,...,p_N]$ trying to resolve the issue, R4P is expected to generate a sequence of tokens, including the reasoning and the IDs of the correct patches.
Then, we can get each patch's predicted correctness (pass or fail). The prompt is formatted:

\begin{table*}[h]
    \centering
    \begin{tabular}{l}
    \toprule
    You are a software expert. You will be given a software issue and some patch \\
    candidates in user query. You need to judge which patch(es) can resolve the issue. \\
    Carefully review, critic, and compare the given candidates. You need to first think \\
    about the reasoning process in the mind until you get the final answer. Finally, put the \\
    ID(s) of correct patch candidates in \texttt{\textbackslash boxed\{\}}. \\
    \\
    \texttt{[ISSUE]} \{user issue $I$\} \texttt{[/ISSUE]} \\
    \texttt{[PATCH 1]} \{agent patch $P_1$\} \texttt{[/PATCH 1]} \\
    $\cdots$ \\
    \texttt{[PATCH N]} \{agent patch $P_N$\} \texttt{[/PATCH N]}\\
    \bottomrule
    \end{tabular}
    \label{tab:prompt}
\end{table*}

This group-level verification fully leverages the non-unique and diverse characteristics of patches, since the models tend to modify various positions with different code changes, where each patch provides sufficient information for judging other patches. As a result, R4P could cross-reference all patches' edit locations and content, inferring potential context and identifying subtle errors in one patch by observing others.
Furthermore, this group-wise approach transforms the binary outcome space into a much denser one (e.g., $\sum C_N^i$ possibilities for random selecting correct patches from a group $P=[p_1,...,p_N]$), which provides a richer supervision signal and significantly mitigates the reward hacking risk inherent in simple binary classification tasks. Specifically, during RL training, the reward of R4P $r(P)$ is calculated by the accuracy of the patch group, where exact match indicator $\mathbb{I}_{y_i = y^*(p_i)}$ equals 1 if the prediction is correct $y_i=y^*(p_i)$ else 0 (Eq.~\ref{eq:reward}).







\textbf{Reward modeling}
To avoid unstable training caused by the model’s random guesses on a patch’s binary correctness when uncertain, we adopt a continuous group-wise reward requiring the model to verify as many correct patches as possible (i.e., group-wise accuracy). The intuition is: a reasoning that correctly verifies more patches should be considered better than one that verifies fewer, even though certain random guessing is inevitable. Thus, the reward clearly reflect how much better one reasoning result is compared to another, rather than treating real correct reasoning as equivalent to a randomly guessed correct answer.
Note that when the number of patches to be verified ($N_v$) exceeds the group size ($N_t$), the set of patches is partitioned into smaller groups for verification, with the total number of inference calls being $\lceil N_v / N_t \rceil$. Conversely, if the number of patches is less than $N_t$, the group is padded with empty patches to match the input format used during training.

\begin{equation}\label{eq:reward}
r(P) =
\begin{cases}
\tfrac{1}{N} \sum_{i=1}^N \mathbb{I}_{y_i = y^*(p_i)} & \text{if boxed} \\
0 & \text{otherwise}
\end{cases}
\end{equation}

\textbf{Data Sampling} To collect a dataset of patches that closely resemble those generated by contemporary SWE agents, we use Claude-3.7-sonnet on OpenHands~\citep{wang2024openhands} scaffold to resolve issues in SWE-Gym~\citep{pan2024training}. For each of the 2,438 issue instances in SWE-Gym, we sampled 6 patch candidates, resulting in a total of 14,628 verified patches. All the labels (pass or fail) of samples are verified by SWE-Gym's sandbox unit testing.
Since the patch pass rate of this process is only 30\%, the original dataset was highly imbalanced with a majority of incorrect patches. Thus, we filtered a subset of the incorrect patches to create a more balanced label distribution. This data is then used for RL training of R4P via GRPO~\citep{guo2025deepseek} (Appendix~\ref{app:r4p}).

\section{Mini-SE}\label{sec:agent}


To validate R4P's practicality, we aim to train an agent where all supervision comes from R4P.
We do not expect R4P to match high-quality execution-based rewards; instead, we want to verify that its scalable yet imperfect supervision can yield stable training gains.
Specifically, most existing agents~\citep{yang2024swe,wang2024openhands} follow a ``generate-then-verify'' workflow that relies on sandbox execution for self-testing during rollout~\citep{jain2025r2e}, coupling generation with verification process. 
Since R4P is scaffold-agnostic, we design Mini-SE, a lightweight, fully execution-free scaffold to cleanly decouple the two process by delegating all verification to R4P.
Its simplicity also avoids the engineering overhead of hosting sandboxes for parallel execution during training and accelerate the process.
To further accelerate the rollout process and reduce the interaction turns caused by long-term, general-purpose tool calls,
Mini-SE adopts an issue-resolution-oriented tool design:


\begin{figure*}[h]
    \centering
    \includegraphics[width=1.0\linewidth]{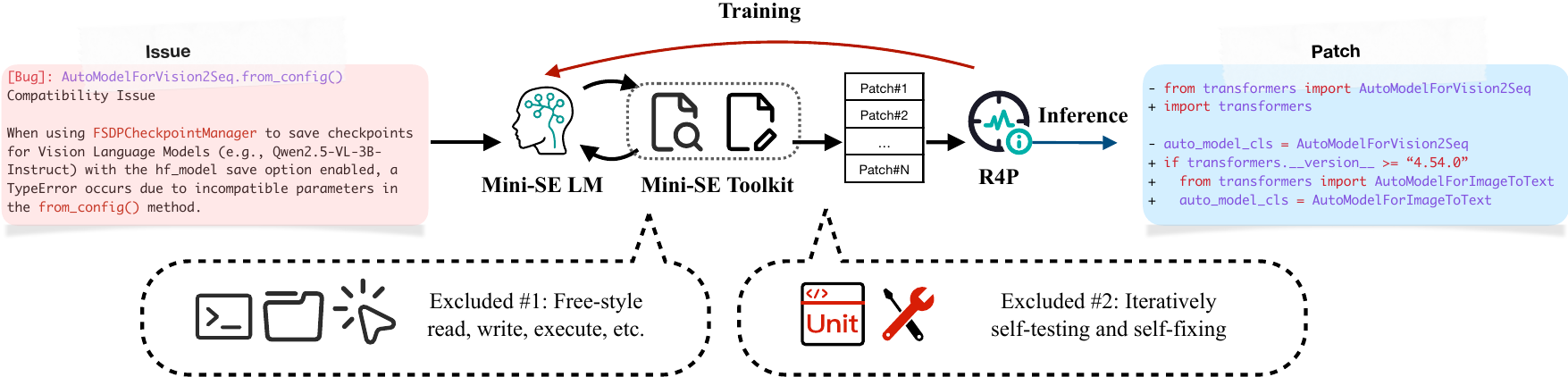}
    \caption{{Mini-SE: a fully execution-free scaffold designed to demonstrate R4P's scalable supervision practicality on downstream tasks.} All patch review burden is shifted to R4P.
    }
    \label{fig:agent}
\end{figure*}

\textbf{Code Search:} A tool that maps an input entity's simple name to corresponding source code and file path.
Upon issue input, it initializes a code graph database for the entire repository, extracting all \textit{class}, \textit{method}, and \textit{function} entities with static analyzer.
To reduce the interaction turns, this tool returns the source code of \textit{all} entities with that simple name (e.g., \texttt{func}).
This \textit{fuzzy matching} also reduces tool call failures due to incorrectly spelled fully qualified names (e.g., \texttt{file.Class.func}).


\textbf{Code Edit:} A tool that replaces an old string with a new string in the input file path, which could work as \textit{insert}, \textit{delete}, and \textit{modify} actions.
Upon issue input, it checkouts the specific commit and creates a shallow clone of the target repository as a workspace.
All modifications are validated using \textit{syntax checker}, and any change with syntax errors is \textit{discarded}. This check reduces failures from the LLM repeatedly trying to fix syntax errors introduced in previous wrong fixes.


Mini-SE focuses solely on issue resolution, generally following a targeted \textit{``entity $\rightarrow$ code $\rightarrow$ patch"} paradigm.
Specifically, it begins with ``symptom" entities mentioned in the issue description, examines its code, and iteratively traces through the dependency chain to identify the ``root cause" entities. Once located, it modifies the corresponding code and submits a patch.
The absence of trial-and-error interactions due to incorrectly using general-purpose tools enables Mini-SE swiftly generating a large volume of candidate patches.
\rev{Appendix~\ref{app:scaffold} details Mini-SE's architecture and contrasts it with existing scaffolds.}


\section{Experiment}

\subsection{Experimental Setup}\label{sec:setup}

\textbf{Implementation} We implement R4P using \texttt{Qwen-2.5-Coder-Instruct-32B} with group size N = 4, as it offers a balanced trade-off between patch verification capability and computational cost. We do not adopt Qwen-3 series for R4P because enabling its thinking mode results in excessively long reasoning chains (around 32K tokens), which significantly increases computational overhead during both training and inference. Disabling thinking mode, however, leads to substantial performance degradation. In contrast, for Mini-SE, we base it on \texttt{Qwen-3-32B} because the agentic interaction pattern involves large volumes of tool-returned tokens rather than model-generated ones, making its training more efficient than that of R4P. Additionally, it demonstrates better tool-use instruction compliance compared to Qwen 2.5 series.
Mini-SE's training details are in Appendix~\ref{app:minise}.



\textbf{Datasets} We collect patches generated by \modify{four different agents (SWE-agent and OpenHands) and LLMs (e.g., gpt-4o) (Appendix Table~\ref{tab:verification_accuracy})} from submissions in the officially verified SWE-bench-verified experiments, with a resolution rate of around 50\% to ensure label balance (Appendix~\ref{app:evaldata}). \edit{None of them shares the same setup used for R4P's training data sampling (Sec.~\ref{sec:r4p}).}
To maintain consistent task difficulty, all empty patches are removed. The final dataset comprises 1,340 patches, forming 335 group-wise instances.
To further assess the practicality of R4P, we use real-world issues from R2E-Gym-Subset~\cite{jain2025r2e} along with rewards generated by R4P to train Mini-SE via RL without using its tests.
\modify{We opt not to use real in-the-wild issues for agent's RL as it introduces many confounding factors beyond test quality, e.g., whether issue-PRs are connected and aligned, whether issues are clear, etc. Thus, we choose to control these factors for fair validation.}
After that, we evaluate Mini-SE on SWE-bench-verified, which consists of 500 issues with testing.

\modify{\textbf{Baselines}: We compare (1) general LLMs (2) trajectory verifiers (3) reward models. For general LLMs we use the same group-wise formulation with R4P; for trajectory verifiers we use their native point-wise formulation with their official controlled \texttt{YES/NO} decoding; for reward models we use a pair-wise formulation. \textit{All baselines are therefore evaluated in their original formats.}}

{
\textbf{Evaluation rationale.}
We evaluate R4P from three perspectives:
\textbf{(1) Direct comparison} (Table~\ref{table:exp_main}, Fig.~\ref{fig:comp}): measuring R4P's patch verification accuracy against general LLMs, trajectory verifiers, and reward models.
\textbf{(2) Downstream practicality} (Fig.~\ref{fig:scaling}): whether R4P's supervision can support stable RL convergence and yield consistent training gains despite its imperfect accuracy (< 100\%), as well as on test-time patch selection.
\textbf{(3) Generalization capability} (Fig.~\ref{fig:scaling}):
{we also compare R4P with all verifiers on DeepSWE agent's trajectory \& patches to examine whether their relative performance is consistent across agents.}
}


\subsection{Main Results}

\begin{figure*}[tbp]
    \centering
    \begin{minipage}{0.4\textwidth}
        \centering
        \captionof{table}{Comparison with general models.\\ (``$*$" mark means point-wise verification)}
        \label{table:exp_main}
        \begin{tabular}{lccc}
            \toprule
            \textbf{Model} & \textbf{Acc.} & \textbf{F1} & \textbf{EM} \\
            \midrule
        claude-3.7-sonnet & 68.1 & 50.5 & 26.8 \\
        claude-4-sonnet & 68.4 & 50.0 & 25.7 \\
        claude-4.5-sonnet & 71.6 & 62.4 & 30.4 \\
        gemini-2.5-pro & 72.7 & 56.6 & 34.6 \\
        gemini-3.1-pro & 73.8 & 56.5 & 31.9 \\
        gpt-4o & 61.2 & 43.0 & 17.3 \\
        gpt-5 & \textbf{76.2} & 62.4 & 41.5 \\
        o4-mini & 68.5 & 52.0 & 29.6 \\
        o3 & 71.5 & 57.4 & 36.4 \\
            \midrule
             qwen-2.5-coder-32b* & 55.9 & - & - \\
            qwen-2.5-coder-32b & 61.9 & 44.5 & 18.5 \\
            \midrule
            R4P\rev{-32b} & 72.2 & \textbf{63.3} & \textbf{41.8} \\
            \bottomrule
        \end{tabular}
    \end{minipage}
    \hfill
    \begin{minipage}{0.5\textwidth}
        \centering
        \includegraphics[scale=0.48]{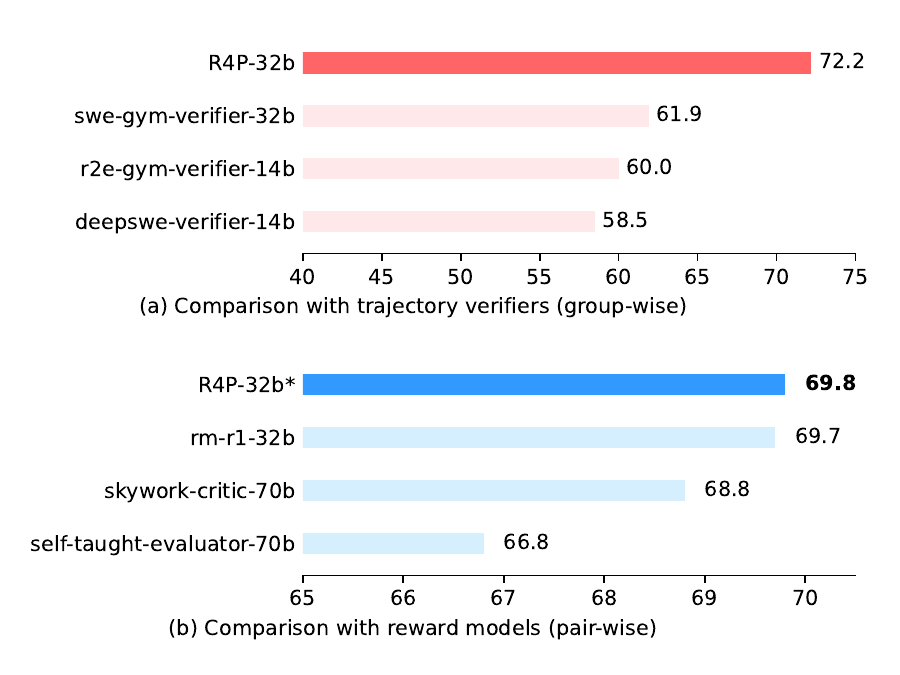}
        \caption{Comparison with RMs. (Acc.)\\ (``$*$" mark means pair-wise selection)}
        \label{fig:comp}
    \end{minipage}
\end{figure*}

\textbf{Direct comparison}
As detailed in Table~\ref{table:exp_main}, we compare R4P with various advanced generative and reasoning LLMs \textit{with the same group-wise formulation}. R4P achieves a patch-level accuracy (Acc.) of 72.2\%, \modify{remaining competitive with proprietary models}, despite R4P having a significantly smaller parameter scale.
Since group-wise verification is a multi-choice retrieval in each group, we also report the group-level F1-score (F1) and Exact Match (EM) {(detailed metric definitions are in Appendix~\ref{app:evaldata}.)}. On these metrics, R4P achieved 63.3\% F1 and 41.8\% EM, outperforming all baselines.
We also compare R4P with trajectory verifiers and general reward models.
For trajectory verifiers, since they can only perform point-wise verification with trajectories, we can only compare their patch-level Acc. with R4P.
As shown in Fig.~\ref{fig:comp} (a), R4P largely outperforms trajectory verifiers, \modify{as it provides a scaffold-ignorant verification on the mixed-scaffold patches and }is much more generalizable. For reward models, since they can only provide relative preference on answers, we construct a pair-wise subset from our evaluation data, where each instance contains one correct and one incorrect patch from the same issue. While this formulation is unoptimized for R4P, it still achieves a 69.8\% accuracy, slightly outperforming the state-of-the-art, as shown in Fig.~\ref{fig:comp} (b). This result underscores that R4P's core verification capabilities are robust and can be effectively generalized to other patch evaluation scenarios.


\begin{figure*}[h]
    \centering
    \begin{minipage}{0.49\textwidth}
        \centering
        \includegraphics[scale=0.45]{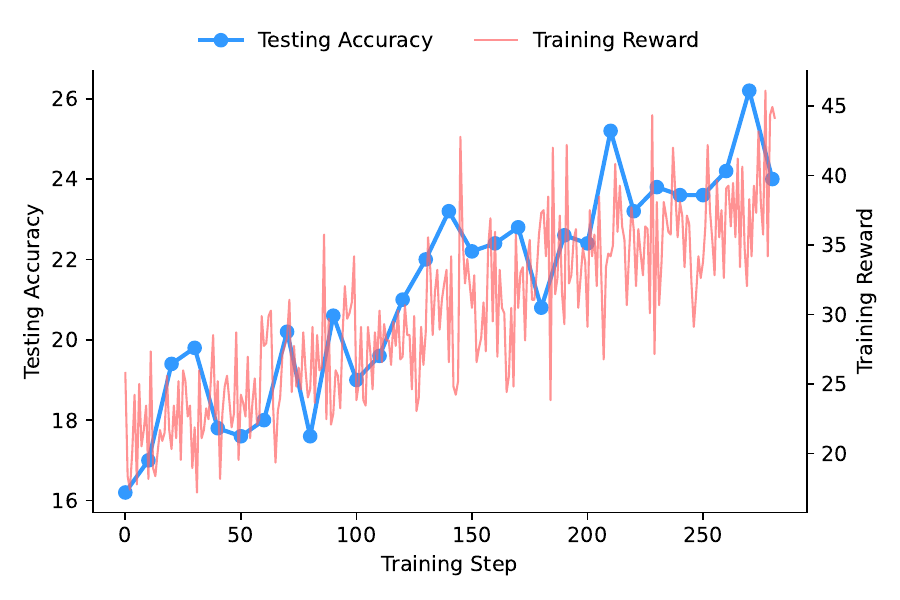}
    \end{minipage}
    \hfill
    \begin{minipage}{0.49\textwidth}
        \centering
        \includegraphics[scale=0.45]{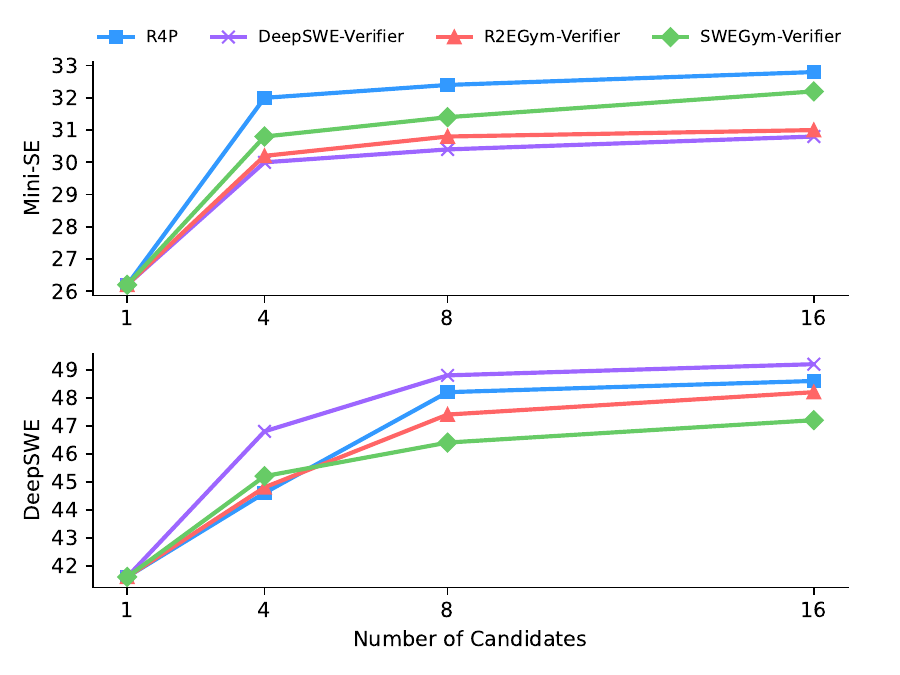}
    \end{minipage}
    \caption{
    \textbf{Left:} (RL practicality): R4P reward leads to a stable RL convergence. \textbf{Upper Right:} (TTS practicality): Resolution rates scale with more patch candidates. \textbf{Lower Right:} (Generalization) R4P is comparable to DeepSWE's verifier on DeepSWE's in-distribution trajectories while others perform worse in out-of-distribution settings.}
    \label{fig:scaling}
\end{figure*}

\textbf{Practicality on RL}
To validate if R4P's supervision could support RL scaling, we trained Mini-SE via \textit{pure RL}.
Notably, unlike normal SWE agents, Mini-SE is lightweight and does not test its patches during rollout. Thus, it naturally achieves a lower Pass@1 score in the absence of an extra test-time patch verifier.
Nevertheless, as shown in Fig.~\ref{fig:scaling},
Mini-SE achieved a 26.2\% Pass@1 within 300 steps, boosting the base Qwen3-32B model performance by 10.0\%.
Furthermore, the test accuracy increases steadily with training rewards, indicating that R4P provides stable supervision signals without leading to reward collapse.
This feature enables R4P to further supervise agents on vast untested data in open-source community, despite its imperfect verification may not be as good as well-designed testing like R2E-Gym's official test suites.
More discussions about sandbox testing and model-based rewarding methods (e.g., R4P) comparison are detailed in Sec.~\ref{sec:analysis}.

\textbf{Practicality on TTS} 
To validate R4P's effectiveness for TTS, we sample 16 patches from Mini-SE on the SWE-bench-verified.
We employed a two-round process: the patches are grouped for R4P verification, and those predicted as incorrect patches will be filtered out. Then we prompt R4P to output the single most likely correct patch from all remaining candidates as output.
As shown in Fig.~\ref{fig:scaling}, Mini-SE's resolution rate increases steadily with the number of candidate patches.
When selected from 16 patches, its score reaches 32.8\%.
{Since existing verifiers are coupled with specific agent's trajectory, R4P can surpass them without relying on additional agent interaction information of Mini-SE.}

\begin{figure*}[h]
    \centering
    \begin{minipage}{0.49\textwidth}
    \centering
    \includegraphics[scale=0.4]{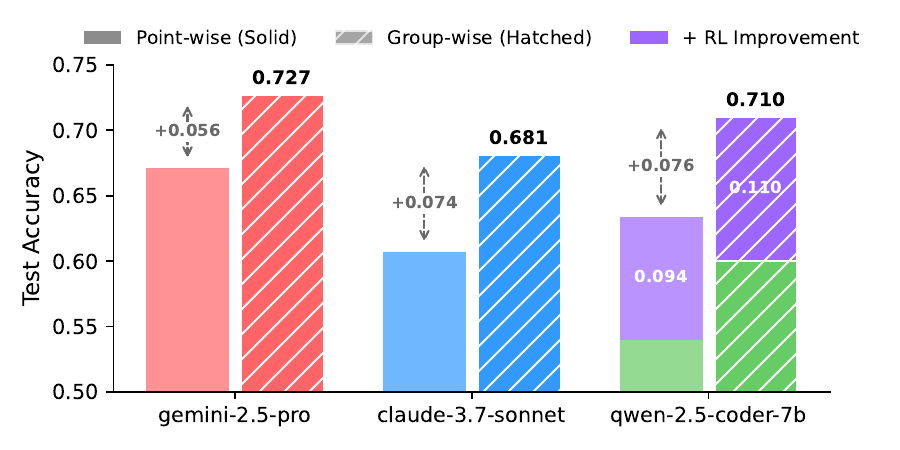}
    \caption{Task formulation}
    \label{fig:comp_model}
    \end{minipage}%
    \hfill
    \begin{minipage}{0.49\textwidth}
    \centering
    \includegraphics[scale=0.4]{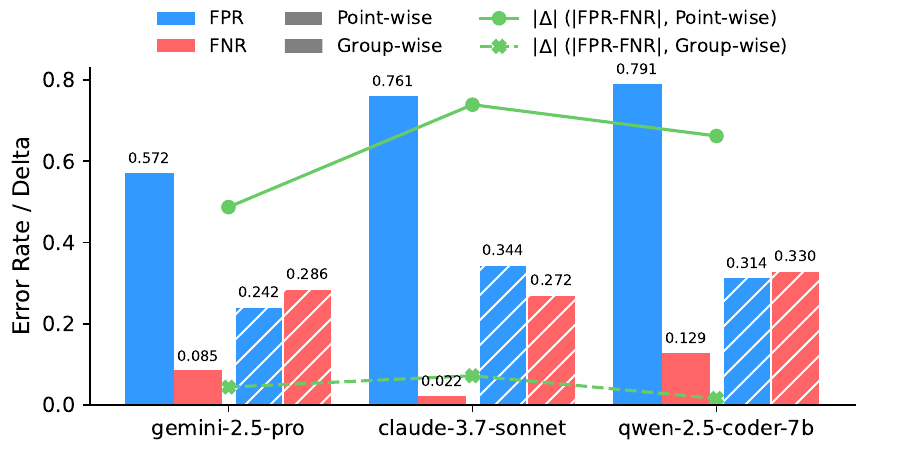}
    \caption{Bias of FPR/FNR}
    \label{fig:bias}
    \end{minipage}%
\end{figure*}

\begin{figure*}[h]
    \begin{minipage}{0.49\textwidth}
    \centering
    \includegraphics[scale=0.4]{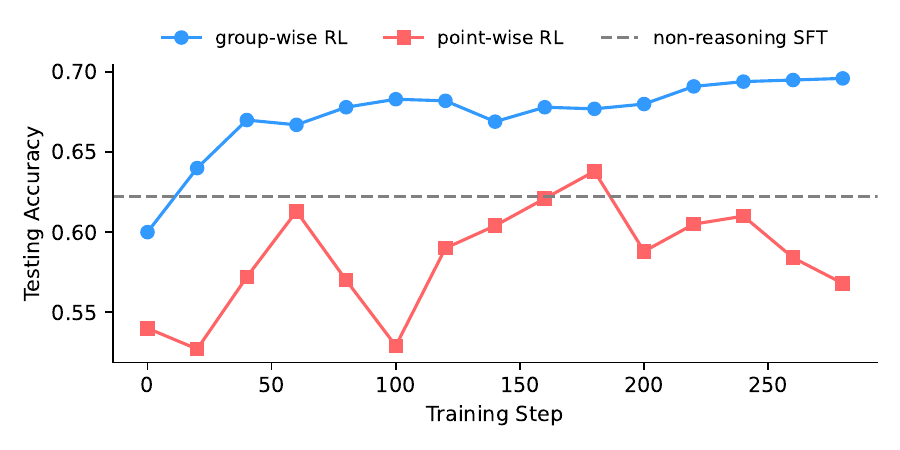}
    \caption{RL convergence}
    \label{fig:comp_step}
    \end{minipage}%
    \centering
    \begin{minipage}{0.49\textwidth}
    \centering
    \includegraphics[scale=0.4]{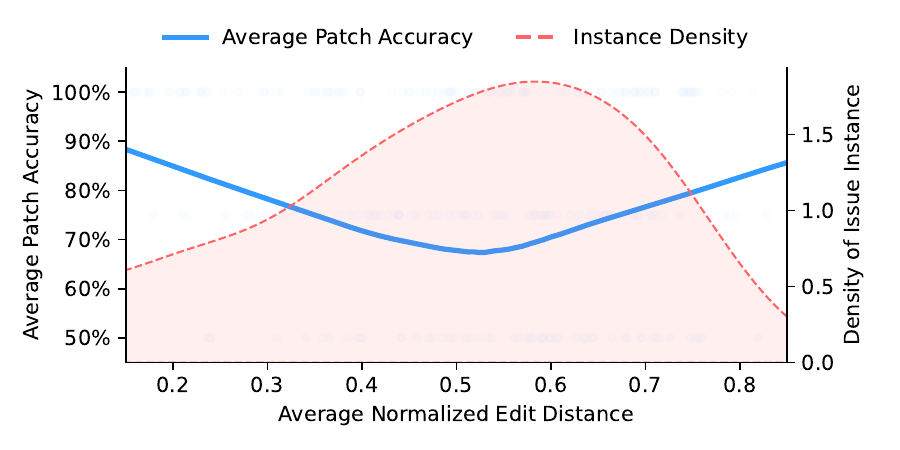}
    \caption{Accuracy of group diversity (edit distance)}
    \label{fig:edit}
    \end{minipage}%
\end{figure*}

\begin{figure*}[h]
    \begin{minipage}{0.49\textwidth}
    \centering
    \includegraphics[scale=0.4]{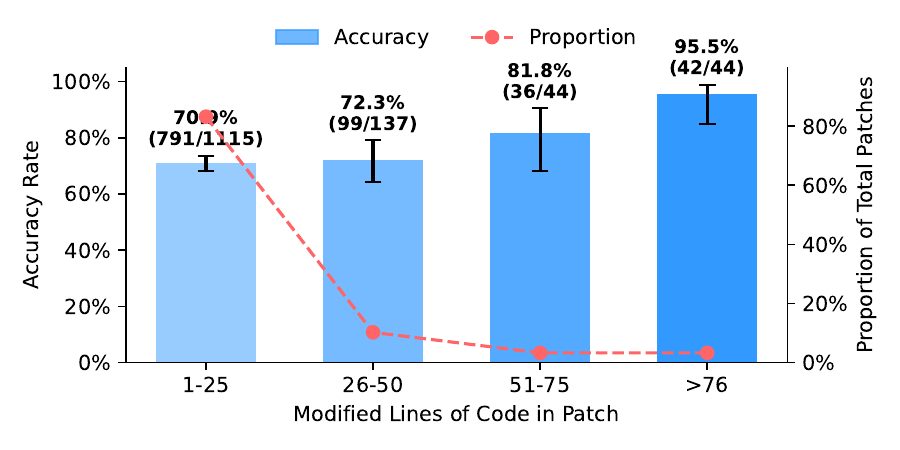}
    \caption{Relations of Acc. and edited lines}
    \label{fig:diffs}
    \end{minipage}%
    \hfill
    \begin{minipage}{0.49\textwidth}
    \centering
    \includegraphics[scale=0.4]{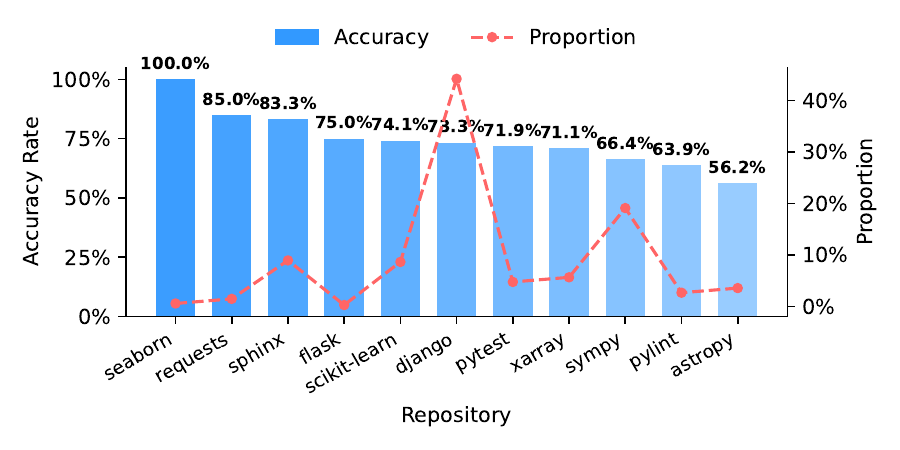}
    \caption{Acc. across repositories}
    \label{fig:repo}
    \end{minipage}%
\end{figure*}

\textbf{Generalization capability}
To further show its generalizability, we employ R4P on trajectories and patches generated by DeepSWE agent.
{
While DeepSWE’s verifier is trained on in-distribution trajectories, R4P still achieves comparable performance without using any trajectory information.
Furthermore, other trajectory verifiers that perform well on Mini-SE show degraded performance on DeepSWE's out-of-distribution trajectories, whereas R4P remains consistently effective across the agents.
These results show that R4P's scaffold-agnostic verification generalizes across different agents.}

\subsection{Analysis and Discussion}\label{sec:analysis}

\textbf{Sandbox testing vs. Model-based rewarding}
The model-based reward is unavoidable contains noise.
Thus, the use of model-based rewards should be based on the premise that \textit{the benefits from data and supervision scaling will outweigh the losses from noisy signals}.
Empirically, the stable reward curve in Fig.~\ref{fig:scaling} shows that R4P's reward is overall stable and will not trigger reward collapse.
Furthermore, we observe that the gradient norm of Mini-SE stabilizes around 0.01.
Compared to existing RL-based agents~\citep{deepswe2025}, Mini-SE’s gradient norm is approximately 25\% higher.
This suggests that although some noisy supervision creates conflicting gradients, the overall update direction remains correct. Given R4P's ability to scale data efficiently, this
{noise is a reasonable cost for scalable, execution-free supervision}.


\textbf{Impact of group-wise patch reasoning.}
Verifying a patch in isolation is a significant challenge for LLMs. As shown in Fig.~\ref{fig:comp_model}, even advanced models like Gemini struggle with this task, while smaller models like Qwen-7B perform at chance level and do not improve much even after targeted RL. However, with group-wise formulation, the performance is improved across all models. With RL, it can be further improved by 10\%.

We identify this improvement stems from two aspects. First, the group-wise reasoning mitigates the tendency of LLMs to \textit{over-estimate} a patch's quality when evaluated in isolation. As shown in Fig.~\ref{fig:bias}, point-wise formulation shows a high false-positive rate, confirming that models inherently struggle to identify errors without sufficient context.
In contrast, group-wise formulation enables LLMs to cross-reference multiple candidate patches to identify the ignorable subtle bugs:
(1) When multiple patches compete to modify adjacent code snippets, the correct one could be distinct against the incorrect ones (e.g., ``\texttt{v=dict.value}" compares to ``\texttt{v=dict.get(value,None)}"); (2) When they modify different code snippets, they provide mutual context that a single patch lacks. As shown in Fig.~\ref{fig:edit}, accuracy approaches 90\% when patches in a group are either very similar or very different, indicating that both conditions simplify the verification.
Therefore, it mitigates the threat of over-estimate. Second, the group-wise formulation resolves the issue of unstable RL due to the sparse outcome space of point-wise verification. As Fig.~\ref{fig:comp_step} illustrates, group-wise formulation enables smooth convergence, where point-wise RL training is highly unstable, and its peak performance barely surpasses that of the non-reasoning SFT.







\textbf{Application scope}
We analyze R4P's application scope to show its comparative advantages and disadvantages.
As shown in Fig.~\ref{fig:diffs}, R4P’s verification accuracy generally appears to be positively correlated with the number of edited lines in a patch, as it cross-references patches' edition content and position as context for effective patch verification.
In addition, Fig.~\ref{fig:repo} demonstrates R4P's robust performance across different repositories, highlighting its generalizability.
Furthermore, as discussed in Fig.~\ref{fig:edit}, R4P's accuracy improves when patches within a group are either highly similar or highly dissimilar. Thus, we suggest several best practices for R4P application: (1) prioritize issues with shorter resolution periods (which is accessible via GitHub API), (2) sampling redundant rollouts and reject rewarding on extremely short patches, and (3) assemble patch groups with either very high/low mutual similarity rather than uniformly distributed for verification.

\edit{Additional analyses on reward consistency and inference efficiency are reported in Appendix~\ref{app:extra_analysis}.}

\section{Related Work}

\textbf{SWE data scaling}
Data scaling has become a central challenge of SWE. Typically, people mines issues from GitHub and creating sandbox for testing~\citep{pan2024training}, which is very labor-intensive. To facilitate it, one way is LLM-assisted sandbox construction~\citep{hu2025llm,yang2025swe}.
However, effort is still required for checking sandbox correctness~\citep{hu2025llm}.
Another way is issue generation and commit backtranslation to mutate numerous instances from several seed instances~\citep{jain2025r2e,yang2025swe}. However, they often differ from real-world issues and empirically lead to suboptimal training performance~\citep{deepswe2025}.
A third approach relies on similarity metrics to supervise SWE sub-tasks separately~\cite{ma2025sorft,wei2025swe,xie2025swe}.
However, since correct patches are often non-unique, it can incorrectly penalize novel yet valid solutions, introducing noise and reducing sample efficiency.
In contrast to these data-centric methods, we focus on expanding supervision rather than data, as a large number of diverse untested issues remain under-exploited in the whole open-source community.
\rev{Learned environment simulation~\citep{sun2026swe} is also orthogonal to R4P: it substitutes the \textit{environment} an agent executes with, whereas R4P frees the \textit{supervision} from execution.}


\textbf{Reward models and verifiers}
Reward models are initially designed to predict an answer's relative quality~\citep{chen2025rm} for AI-human preference alignment.
With the emergence of Reinforcement Learning with Verifiable Rewards (RLVR) and Monte Carlo-based RL algorithms, e.g.,  GRPO~\citep{guo2025deepseek,shao2024deepseekmath}, rule-based rewards have gained prominence in reasoning tasks where absolute correctness can be determined. In such settings, reward models are often repurposed as verifiers during test-time selection~\citep{pan2024training,jain2025r2e,deepswe2025}. However, SWE tasks lack easily accessible high-quality rule-based rewards from testing.
Thus, we explore using reward models with absolute patch verification capability to provide scalable, deterministic supervision.

\section{\edit{Conclusion}}

This work explores a model-based approach to address the scalability bottleneck in supervising software engineering agents due to reliance on testing. It introduces R4P, a reasoning-based patch verification method. R4P utilizes group-wise training objective to effectively verify patches against each other and mitigate instability during training its patch reasoning capability.
Empirical results demonstrate that R4P achieves 72.2\% verification accuracy, \edit{competitive with frontier proprietary models like Claude-4.5-Sonnet, GPT-5, and Gemini-3.1-Pro}. We further showcased its practical value by training Mini-SE, a fully \edit{execution-free} agent that leverages R4P for both reinforcement learning and test-time scaling. Mini-SE achieves 26.2\% Pass@1 and 32.8\% TTS (N=16) on SWE-bench-verified, showing that reasoning-based supervision
{can unlock the potential of data scaling by providing usable supervision where traditional test suites are unavailable}.

\section*{\edit{Limitations}}

\textit{First}, R4P's weights remain fixed after training. \modify{Similar to all reward models,} as the agent's policy improves, the static reward model may become misaligned with true answer quality~\citep{gao2023scaling}. In future work, we plan to explore periodic calibration using issues with high-quality sandbox during extended RL training, where discrepancies between R4P's predictions and actual test results could guide weight updates.
\textit{Second},
{while R4P's patch verification is trajectory-free and scaffold-agnostic, it cannot supervise an agent's in-loop self-testing and self-fixing behaviors, which require execution environments~\citep{gao2025trae,deepswe2025}.} Future work will explore mixed training strategies, e.g., scaling normal issues with R4P to train straightforward patch generation while leveraging challenging issues with high-quality test sandbox to train patch generation with in-loop verification.
\textit{Third}, R4P and Mini-SE are Python-centric. To generalize to other languages, it is required to collect patches from their datasets and adapt the code search tool to their static analyzer.
{\textit{Fourth}, our current experiments validate R4P on existing datasets, as correctly connecting issues with corresponding PRs in less-maintained codebases remains a challenge. Future work will explore collecting and training on vast untested in-the-wild issues.}
\textit{Fifth}, R4P's performance correlates with multiple factors like group diversity as discussed in Sec.~\ref{sec:analysis}. To mitigate it, it might be helpful to follow our best practice suggestions.

\section*{\edit{Ethics Statement}}
\edit{This paper presents work whose goal is to advance the field of Machine Learning and Software Engineering in the domain of AI coding. This work uses only publicly available datasets (SWE-Bench, SWE-Gym, R2E-Gym) and their corresponding public submissions (SWE-Bench Experiments), involving no human subjects or sensitive personal data. These datasets are released for research use under permissive open-source licenses by their respective authors, and our use is consistent with their intended research purpose; we do not redistribute the underlying data. The data consists of public GitHub pull requests aggregated by the original benchmark curators, and we do not perform additional personally identifying information (PII) filtering beyond what these source datasets already apply. Derivative artifacts of R4P or Mini-SE should not be deployed or used in real-world software systems without a comprehensive security evaluation, since model-based patch verification can mark insecure or backdoored patches as acceptable. We have considered potential ethical issues including fairness, privacy, and security, and found no concerns arising from our methodology or results.}

\section*{\edit{Use of AI Assistance}}
\edit{We used general-purpose large language models to polish language and to check LaTeX formatting. All research ideas, experimental design, analysis, and conclusions are the authors' own, and the authors take full responsibility for the content of this paper.}

\section*{Acknowledgements}
This paper was supported by the Guangdong Basic and Applied Basic Research Foundation (No.2024A1515010145) and the Shenzhen Science and Technology Program (Shenzhen Key Laboratory Grant No. ZDSYS20230626091302006).

\bibliography{ref}

\newpage
\appendix
\section{Preliminary Study of Tested Issues}\label{app:pre}
\begin{figure*}[htbp]
    \centering
    \includegraphics[width=0.9\linewidth]{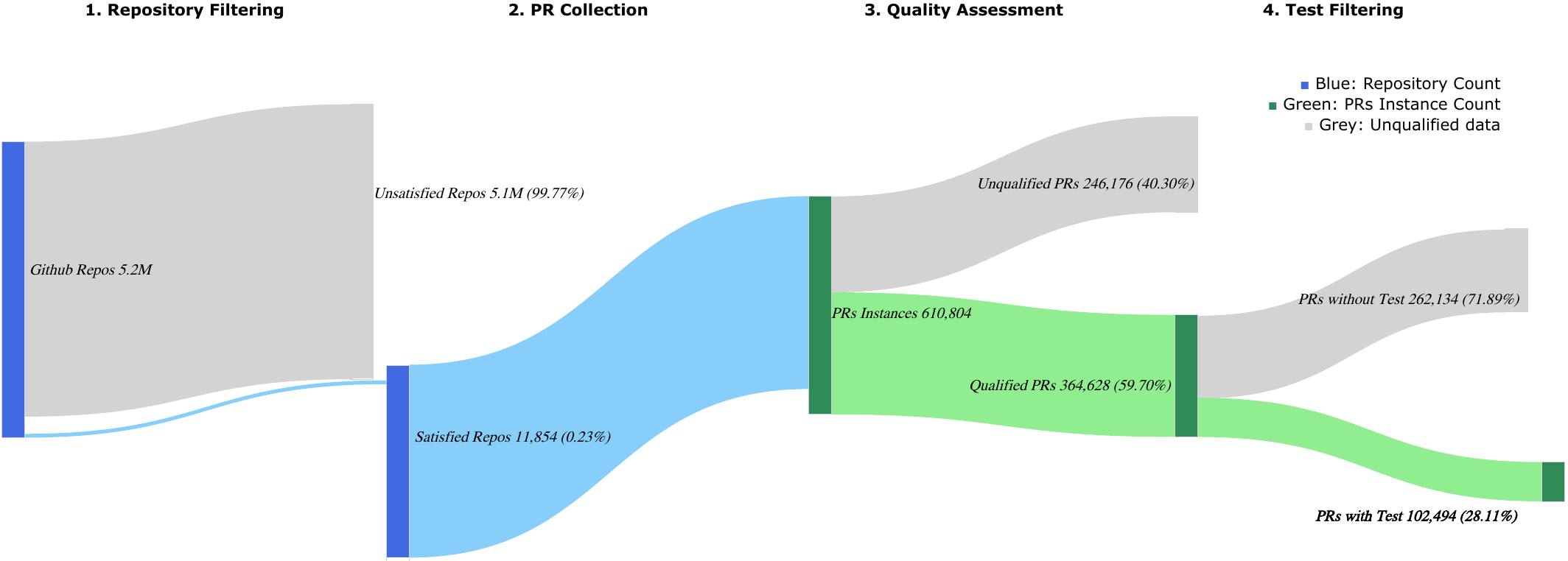}
    \caption{Visualization of data flow across four hierarchical steps in our GitHub issue curation pipeline: (1) \emph{Repository Filtering}, (2) \emph{PRs Collection}, (3) \emph{Quality Assessment}, and (4) \emph{Test Filtering}. The text in \textbf{\textcolor{blue!70}{Blue}} indicates repository counts, the text in \textbf{\textcolor{green}{Green}} indicates issue–instance counts, and \textbf{\textcolor{gray!30}{Grey}} flows denote data filtered out as unsatisfied or unqualified at each stage. Numbers and percentages shown in the diagram are computed within each step.}
    \label{fig:data_collection_workflow}
\end{figure*}

This study measures the prevalence of associated tests for issues and pull requests in open-source communities.
\noindent We begin from the universe of 5.2 M public GitHub repositories and pre-filter to Python projects with at least 500 stars, excluding any repository already included in \textit{SWE-Bench Verified}. The repository filtering leaves 11,854 eligible projects (0.23\%). From these repositories, we harvest 610,804 PRs. We then apply a series of quality filters to the candidate issues: (i) the problem description must fall within 100--4{,}000 characters to align with the length distribution observed in \textit{SWE-Bench Verified}; (ii) descriptions containing non-ASCII content (e.g., Chinese characters) are discarded; (iii) associated fixes must not modify non-Python files; (iv) patches must be small and self-contained, touching 1--5 files; and (v) issues embedding images in the description are removed to avoid non-textual ambiguity. Finally, in qualified PRs, we filter the patch without corresponding test changes, leaving only 102,494 qualified PRs with tests. As summarized in Fig.~\ref{fig:data_collection_workflow}, this pipeline yields only \textbf{28.11\%} PRs have corresponding tests.

\section{Experiment Details}

\subsection{Training R4P}\label{app:r4p}

\paragraph{Training settings}
We adopt GRPO algorithm for training R4P. Based on the insights from recent DAPO report, we set the clip ratio high to 0.28 and do not employ a KL loss in objective function. With a batch size of 128, we trained Qwen-2.5-Coder-Instruct for 225 steps.
To accelerate the convergence process, R4P generates 16 samples and performs 16 gradient backward in each step.
Throughout training, the temperature is maintained at 1.0 and the learning rate is fixed at 1e-6.
To implement the training, we adopt VeRL framework with vLLM engine for rollout.
\edit{R4P training was conducted on 32$\times$NVIDIA H20 GPUs.}

\paragraph{Dataset construction}
As mentioned in Sec.~\ref{sec:r4p}, we sampled 6 patches for each issue in the SWE-gym sandbox. During training data construction, the group size was set to 4. To improve the sample efficiency of the patches, we created multiple instances for each issue by forming different combinations of patches, i.e., constructing $C_6^4$ combinations for each issue. However, since the average solve rate of Claude-3.7 with OpenHands on SWE-gym is only about 30\%, directly use such label imbalanced data will lead to significant under-estimate tendency of LLM.
Thus, we balanced the combinations such that each group contained a roughly equal number of instances with correct patches ranging from 0 to 4—each category accounting for approximately 50\% of the total instances. Ultimately, this process yielded 7,599 training samples with a group size of 4.

\paragraph{Prompt template}
We provide the system and user prompt in our R4P implementation, where the verification group size N=4:

\textit{System prompt}
\begin{lstlisting}
You are a software expert. You will be given a software issue and some patch candidates in user query. You need to judge which patch(es) can resolve the issue. Carefully review, critic, and compare the given candidates. You need to first think about the reasoning process in the mind until you get the final answer. Finally, put the ID(s) of correct patch candidates within \boxed{}, e.g., \boxed{1}, \boxed{2, 4}, \boxed{1, 2, 3, 4} (all correct), \boxed{} (all wrong).
\end{lstlisting}

\textit{User prompt}
\begin{lstlisting}
<issue>
{problem_statement}
</issue>
<patch-1>
{generated_patch_1}
<patch-1>
<patch-2>
{generated_patch_2}
<patch-2>
<patch-3>
{generated_patch_3}
<patch-3>
<patch-4>
{generated_patch_4}
<patch-4>
\end{lstlisting}


\subsection{Training Mini-SE}\label{app:minise}

\paragraph{Training settings}
We mainly follow the best practice provided by DeepSWE, including using leave-one-out and remove reward standard deviation for advantage estimation, filter out overlong trajectory without loss calculation, and normalize policy loss by sequence length. During training, we adopt fully on-policy training, with batch size of 64 and sampling 8 rollout per step. We also fix the temperature at 1.0 and the learning rate at 1e-6. To implement agentic RL, we adopt VeRL's AgentLoop framework for training, and use vLLM for rollout. To accelerate the training process, we extract all instance's code graph via tree-sitter in advance, which can be directly used for code search tool. We also checkout the copy of repositories in a separate directory for models to edit. All the code changes will be recorded in the final patches generated by git diff. In addition, we set a maximum limit of 50 rounds and 28K tokens during training. We deploy R4P model on a vLLM server for providing verification. \modify{The reward for agent's RL training is vanilla rule-based reward (1 for ``correct" patch, 0 for ``incorrect" patch).}
We finish the training process before 300 steps.
\edit{Mini-SE training was conducted on the same 32$\times$NVIDIA H20 GPUs; the combined wall-clock time for R4P and Mini-SE training was approximately 7 days.}

\paragraph{Prompt template} We provide the system prompt and tool configs of Mini-SE for reference. The user prompt is the SWE-bench or R2E-gym problem statement itself.

\textit{System prompt}
\begin{lstlisting}
You are an expert AI software engineering agent. Your primary goal is to resolve a GitHub issue given in the user message. Following this workflow methodically:

1.  Understand the problem:
    - Thoroughly comprehend the issue description, identifying core components and expected behavior
    - Determine reproduction steps and failure conditions
2.  Explore and Locate:
    - Use `search_tool` to explore the relevant files, entities, and test cases related to the bug report
    - Locate the exact root cause of the bug
3.  Develop and Fix:
    - Develop minimal, targeted code modifications to address the root cause
    - Use `edit_tool` to apply surgical patch. Aim for minimal, clean changes
4.  Review and Revise:
    - Review the original reproduction steps to ensure the fix effectiveness
    - Review the relevant regression tests to avoid introducing any new bugs
    - Iterate using `search_tool` for review and `edit_tool` for revise until you confirm no edge cases are overlooked
5.  Submit the patch:
    - Call `patch_submission` tool to generate a unix diff patch and submit it to the user when confirming full resolution
    - Ensure the final patch is non-empty before finishing this conversation
    - All code changes persist throughout the conversation and will be included in the final patch
\end{lstlisting}

\textit{Tool configs}

\begin{lstlisting}
  type: "function"
  function:
    name: "edit_tool"
    description: ``A file edit tool that replaces an old string of text with a new string.\nNotes:\n1.  The `old_str` parameter must match a segment of the file's content **exactly**. Pay close attention to whitespace, indentation, and newlines.\nThe edit will fail if `old_str` is not found, or if it is found multiple times in the file. Ensure `old_str` is unique enough to target the specific code block.\n3. `edit_tool` **permanently modifies the actual repository** (changes persist to the final outcome)."
    parameters:
      type: "object"
      properties:
        path:
          type: "string"
          description: "Relative file path e.g. `dir/file.py`."
        old_str:
          type: "string"
          description: "The exact string/content to be replaced in the file. This must be a unique match within the entire file."
        new_str:
          type: "string"
          description: "The new string that will replace `old_str`. Use an empty string to perform a deletion."
      required: ["path", "old_str", "new_str"]


  type: "function"
  function:
    name: "search_tool"
    description: "A code graph tool to show the source code of a function/class/class_method by its name.\nNotes:\n1. This tool is designed for Python code; it cannot find entities in other languages.\n2. If the `entity` name is not unique across the repository, the source code for all matching items will be displayed.\n3. This tool operates on a static, **initial snapshot of the repository**. Changes made with `edit_tool` will **not** be visible in the search results."
    parameters:
      type: "object"
      properties:
        construct:
          type: "string"
          description: "Type of program construct to search for."
          enum: ["function", "class", "class_method"]
        entity:
          type: "string"
          description: "The simple name of the entity to find (e.g., `my_function`, `MyClass`, `my_class_method`)."
      required: ["construct", "entity"]

  type: "function"
  function:
    name: "patch_submission"
    description: "Automatically generate a diff patch for existing code changes and submit it to the user."
    parameters:
      type: "object"
      properties: {}
      required: []
\end{lstlisting}

Note that we also provide a patch submission tool to enable a final check of LLM before ending the interaction and the real patch submission process.

\subsection{Evaluation Data}\label{app:evaldata}
\paragraph{Data source}
Our evaluation data for patch verification task is sampled from SWE-bench verified experiments, which include all the patch submissions of the agents on the leaderboard. Specifically, our training data is from the following submissions: (1) 20250520\_openhands\_devstral\_small, (2) 20241029\_openhands-codeact-2.1-sonnet-20241022, (3) 20240728\_sweagent\_gpt4o, (4) 20240620\_sweagent\_claude3.5sonnet, as they have a overall moderate accuracy for balancing patch labels.
Note that we remove the empty patches and patches that have meaningless modifications (e.g., only comments or tests are changed) to avoid model shortcut through these low-level features.
We thereby formulate a final datasets with 1,340 patches from 335 issues, where the overall balanced T/F label distribution (49.9\% : 50.1\%).
The detailed data distribution and R4P's performance on each sub-set are as follow:

\begin{table*}[h]
\centering
\caption{Verification accuracy and ground-truth label distribution for different LLM agents.}
\label{tab:verification_accuracy}
\begin{tabular}{lcc}
\toprule
\textbf{LLM Agents} & \textbf{Acc. (\%)} & \textbf{Ground-truth T : F} \\
\midrule
20250520-openhands-devstral-small & 74.03 & 66.6\% : 33.4\% \\
20241029-openhands-codeact-2.1-sonnet-2024102 & 72.83 & 59.4\% : 40.6\% \\
20240728-sweagent-gpt4o & 75.22 & 33.1\% : 66.9\% \\
20240620-sweagent-claude3.5sonnet & 66.87 & 40.6\% : 59.4\% \\
\midrule
{Overall} & {72.23} & {49.9\% : 50.1\%} \\
\bottomrule
\end{tabular}
\end{table*}

\paragraph{Prompt for reward model}
We follow the official prompt template of these pair-wise reward models:

\textit{System prompt}
\begin{lstlisting}
    Please act as an impartial judge and evaluate the quality of the responses provided by two AI assistants to the user question displayed below. You should choose the assistant that follows the user's instructions and answers the user's question better. Your evaluation should consider factors such as the helpfulness, relevance, accuracy, depth, creativity, and level of detail of their responses. Begin your evaluation by comparing the two responses and provide a short explanation. Avoid any position biases and ensure that the order in which the responses were presented does not influence your decision. Do not allow the length of the responses to influence your evaluation. Do not favor certain names of the assistants. Be as objective as possible. After providing your explanation, output your final verdict by strictly following this format: \\"[[A]]\\" if assistant A is better, \\"[[B]]\\" if assistant B is better.
\end{lstlisting}

\textit{User prompt}
\begin{lstlisting}
[User Question]
{issue}


"[The Start of Assistant A's Answer]
{patch_1}
[The End of Assistant A's Answer]


[The Start of Assistant B's Answer]
{patch_2}
[The End of Assistant B's Answer]
\end{lstlisting}

\paragraph{Prompt for trajectory verifier}
We follow the official prompt template of these verifiers:

\textit{System prompt}
\begin{lstlisting}
You are an expert judge evaluating AI assistant interactions. Your task is to determine if the assistant successfully resolved the user's request.

Key evaluation criteria:
1. Did the assistant complete the main task requested by the user?
2. Did the assistant handle all edge cases and requirements specified?
3. Were there any errors or issues in the final solution?
4. Did the assistant verify the solution works as intended?

Respond only with "<judgement>YES</judgement>" or "<judgement>NO</judgement>".
\end{lstlisting}

where the user prompt is specific agent's trajectory.

\paragraph{\modify{Evaluation metrics}}

If a dataset has $N$ patches and $M$ unique issues (where each issue corresponds to $K$ patches), then these $N$ patches can be grouped into $M$ patch groups (each of size $K$, i.e., $N = M \times K$). Below are the detailed definitions:

\begin{itemize}[leftmargin=*]
    \item \textbf{Acc.:} Accuracy is a \textit{patch-level} evaluation metric that quantifies the proportion of correctly verified patches out of the total number. Let $y_i$ be the ground-truth label for patch $p_i$, and $y^*(p_i)$ be the model's prediction of its correctness. Using the indicator function $\mathbb{I}$ (which evaluates to 1 if the condition is true, and 0 otherwise), the accuracy is calculated as:

\begin{equation}
    Acc=\frac{\sum^N_{i=1}\mathbb{I}_{y_i = y^*(p_i)}}{N}
\end{equation}

    \item \textbf{F1} The F1 score is a \textit{group-level} evaluation metric. In our implementation, we calculate it using the set-theoretic definition (i.e., the Dice coefficient). For a given group $i$:

- Let $A_i$ be the set of truly correct patches.

- Let $B_i$ be the set of patches predicted as correct by the model.

The F1 score for group $i$ is calculated as:

\begin{equation}
    F1_i=\frac{2|A_i\cap B_i|}{|A_i|+|B_i|}
\end{equation}

This metric serves to measure the similarity between the predicted set ($B_i$) and the ground-truth set ($A_i$). Furthermore, it can provide an alternative view of a \textit{retrieval task} i.e., \textit{the task of "retrieving the correct patches from a set of candidate patches."} In this context, the F1 score represents the harmonic mean of Precision ($P_i$) and Recall ($R_i$):

\begin{equation}
\begin{aligned}
F1_i &= \frac{2P_iR_i}{P_i+R_i} = \frac{2TP_i}{2TP_i+FP_i+FN_i} \\
     &= \frac{2|A_i\cap B_i|}{|A_i|+|B_i|}
\end{aligned}
\end{equation}

Note that in our implementation, we calculate the overall average F1 score for each individual group:

\begin{equation}
    F1=\frac{\sum^M_{i=1}F1_i}{M}
\end{equation}

To handle the boundary conditions where one or both sets might be empty, we consider:

- $F1_i = 1$ if both sets are empty, as the model correctly predicted that no patch is correct.

- $F1_i = 0$ if only one set is empty, as it is a completely incorrect prediction.

    \item \textbf{EM} Exact Match is a \textit{group-level} metric that assesses the model's ability to correctly verify all patches within an entire group simultaneously. It is formally defined as the proportion of patch groups where the predicted set of correct patches exactly matches the ground-truth set:

\begin{equation}
    EM=\frac{\sum^M_{i=1}\mathbb{I}_{|A_i|=|B_i|}}{M}
\end{equation}

\end{itemize}

\section{\rev{Mini-SE Architecture and Scaffold Comparison}}\label{app:scaffold}

\rev{\textbf{Internal architecture.}
Figure~\ref{fig:minise_arch} depicts Mini-SE's internal structure.
Upon issue input, Mini-SE performs a one-time initialization: it builds a \textit{code graph} that indexes all \textit{class}, \textit{method}, and \textit{function} entities of the repository via static analysis, and creates a shallow-cloned \textit{workspace} checked out at the target commit.
The LM then alternates between two tools in a short, execution-free loop:
\textit{CodeSearch} maps an entity's simple name to the source code and file paths of \textit{all} matching entities in the code graph (fuzzy matching);
\textit{CodeEdit} replaces an old string with a new string in a given file of the workspace, where each modification is validated by a syntax checker and discarded upon errors.
Following the ``entity$\rightarrow$code$\rightarrow$patch'' paradigm, the LM starts from the symptom entities mentioned in the issue, traces dependency chains to locate root-cause entities, edits the corresponding code, and submits the accumulated workspace diff as the final patch.
No test generation, execution, or self-verification occurs in the loop: all verification is delegated to R4P, which rewards patches during RL training and selects patches at test time.}

\begin{figure*}[t]
\centering
\resizebox{0.98\textwidth}{!}{%
\begin{tikzpicture}[black,
  font=\small,
  box/.style={draw, rounded corners, align=center, inner sep=5pt},
  tool/.style={box, fill=black!8, minimum width=4.0cm, minimum height=2.8em},
  data/.style={box, fill=gray!12, minimum width=3.6cm, minimum height=2.8em},
  io/.style={box, fill=orange!15, minimum height=2.8em},
  arr/.style={-{Stealth[length=2mm]}, semithick},
  init/.style={arr, dashed},
  lbl/.style={font=\scriptsize, align=center, inner sep=1.5pt}
]
  \node[io] (issue) {Issue};
  \node[box, fill=green!12, minimum height=3.4cm, minimum width=1.8cm, right=1.2cm of issue] (lm) {Mini-SE\\LM};
  \node[tool, anchor=west] (search) at ([xshift=2.9cm, yshift=1.25cm]lm.east) {\textbf{CodeSearch}\\ \scriptsize entity name $\rightarrow$ code (fuzzy match)};
  \node[tool, anchor=west] (edit) at ([xshift=2.9cm, yshift=-1.25cm]lm.east) {\textbf{CodeEdit}\\ \scriptsize string replace + syntax check};
  \node[data, right=1.3cm of search] (graph) {Code graph\\ \scriptsize class / method / function entities};
  \node[data, right=1.3cm of edit] (ws) {Workspace\\ \scriptsize shallow clone @ target commit};
  \path (graph.east) -- (ws.east) coordinate[midway] (backmid);
  \node[io, right=1.0cm of backmid] (repo) {Repository};
  \node[io, below=1.4cm of lm] (patch) {Patch\\ \scriptsize workspace diff};
  \node[box, fill=red!10] (r4p) at (issue|-patch) {R4P\\ \scriptsize reward (RL) /\\ \scriptsize selection (TTS)};
  \coordinate (s-in) at ([yshift=5pt]search.west);
  \coordinate (s-out) at ([yshift=-5pt]search.west);
  \coordinate (e-in) at ([yshift=5pt]edit.west);
  \coordinate (e-out) at ([yshift=-5pt]edit.west);
  \draw[arr] (issue) -- (lm);
  \draw[arr] (lm.east|-s-in) -- (s-in) node[lbl, midway, above] {entity simple name};
  \draw[arr] (s-out) -- (lm.east|-s-out) node[lbl, midway, below] {code + file paths (all matches)};
  \draw[arr] (lm.east|-e-in) -- (e-in) node[lbl, midway, above] {(file, old str, new str)};
  \draw[arr] (e-out) -- (lm.east|-e-out) node[lbl, midway, below] {diff / discard on syntax error};
  \draw[{Stealth[length=2mm]}-{Stealth[length=2mm]}, semithick] (search) -- (graph) node[lbl, midway, above] {lookup};
  \draw[arr] (edit) -- (ws) node[lbl, midway, above] {apply valid edit};
  \draw[init] (repo.north) |- (graph.east) node[lbl, pos=0.62, above] {static analysis};
  \draw[init] (repo.south) |- (ws.east) node[lbl, pos=0.62, below] {shallow clone};
  \draw[arr] (lm) -- (patch) node[lbl, midway, right] {submit};
  \draw[init] (ws.south) |- (patch.east) node[lbl, pos=0.85, above] {diff};
  \draw[arr] (patch) -- (r4p) node[lbl, midway, above] {verify};
  \node[draw, dashed, rounded corners, inner sep=9pt, fit=(lm)(search)(edit),
        label={[lbl]north:{Execution-free interaction loop: \textit{entity} $\rightarrow$ \textit{code} $\rightarrow$ \textit{patch} (no test generation or execution)}}] {};
\end{tikzpicture}%
}
\caption{\rev{Internal architecture of Mini-SE. Dashed arrows denote one-time initialization upon issue input or the final diff extraction; solid arrows denote the LM's interaction loop. All verification is delegated to R4P, outside the scaffold.}}
\label{fig:minise_arch}
\end{figure*}

\rev{\textbf{Comparison with existing scaffolds.}
Table~\ref{tab:scaffold_comp} contrasts Mini-SE with three representative scaffolds: SWE-Agent~\citep{yang2024swe}, OpenHands~\citep{wang2024openhands}, and DeepSWE~\citep{deepswe2025}.
Mini-SE deliberately trades generality (arbitrary shell commands, in-loop self-testing) for rollout efficiency and infrastructure simplicity: it needs no sandbox hosting, produces short trajectories, and cleanly decouples patch generation from verification, which suits its role as a testbed for validating R4P's scaffold-agnostic supervision.}

\begin{table*}[t]
\centering
\footnotesize
\setlength{\tabcolsep}{4pt}
\caption{\rev{Comparison between Mini-SE and representative SWE-agent scaffolds.}}
\label{tab:scaffold_comp}
\begin{tabular}{@{}p{2.5cm}p{2.4cm}p{2.2cm}p{3.4cm}p{3.0cm}@{}}
\toprule
\textbf{Scaffold} & \textbf{Tool set} & \textbf{Execution access} & \textbf{Verification source} & \textbf{Interaction pattern} \\
\midrule
SWE-Agent~\citep{yang2024swe} & bash, file editor, search, bash & Sandboxed (Docker) & Self-testing via execution in the sandbox & Free-form exploration with trial-and-error \\
\addlinespace
OpenHands~\citep{wang2024openhands} & bash, IPython, browser, file editor & Sandbox (Docker) & In-loop self-testing via execution & Event-stream loop, generate-then-verify \\
\addlinespace
DeepSWE~\citep{deepswe2025} & bash, file editor, search, finish & Sandbox (Docker) & Execution-based RL reward; hybrid TTS verifier & Multi-turn rollout with in-loop self-verification \\
\midrule
Mini-SE (ours) & CodeSearch, CodeEdit & None (execution-free) & R4P reasoning-based reward for RL and TTS & ``entity$\rightarrow$code$\rightarrow$patch'' loop, no self-testing \\
\bottomrule
\end{tabular}
\end{table*}

\edit{\section{Additional Analyses}\label{app:extra_analysis}}

\edit{\textbf{Reward consistency.} We analyzed the distribution of predictions that diverged from the majority vote. When sampling 32 times with temperature=1.0, the results indicate high consistency: 88.8\% of patches had no predictions deviating from the majority, and 3.2\% had only one deviation across all runs. Overall prediction fluctuations are fewer than 8\% of cases. Such stability ensures that our reasoning-based verification provides consistent rewards online, improving the reproducibility of training and reducing the needs for inefficient majority voting on rewarding.}

\edit{\textbf{Efficiency on patch verification.} We compare R4P against the time taken to test the golden patches for 500 instances from SWE-bench-verified using a CPU server with 64 cores and 128GB of RAM. The average time per instance for R4P does not exceed 1 second, significantly outperforming the average testing time per instance of 50 seconds. Furthermore, patches generated by LLMs often contain algorithmic inefficiencies or infinite loops. During the training of Mini-SE, we observed that 59.2\% (29/49) of validation instances reached the 30-minute timeout of SWE-bench harness. Furthermore, when given a limited context token length quota, R4P can verify a group of patches in a single run, much more efficient than current trajectory-based verifiers, as the trajectories are often overlong and needed truncation to enable the valid inference for a patch. This demonstrates the superior efficiency of R4P's group-wise reasoning-based verification.}

\section{\modify{Case Study}}

In this section, we provide a case study of \texttt{django\_\_django-13810}. We release R4P's reasoning trajectory as well as those of proprietary models with accessible reasoning trajectory (Claude-3.7-Sonnet, Claude-4.0-Sonnet) as a reference.

\paragraph{Issue and patches}

This issue is about Django's \textit{MiddlewareNotUsed} error. In the original code, the handler variable is updated via \textit{self.adapt\_method\_mode()} before the middleware instantiation triggers the exception. Thus, even though the middleware is skipped, the handler remains permanently modified (incorrectly adapted for subsequent middleware), leading to a sync/async mismatch.

To solve this issue, each patch has its own way:

\begin{itemize} 
\item \textbf{Patch 1 \& 4} (incorrect): Attempt to initialize the middleware \textit{before} adapting the handler. This fails because middleware constructors typically expect the handler to already be in the specific mode (sync/async) required.
\item \textbf{Patch 2} (correct): Assigns the adapted handler to a temporary variable. The main handler is only updated in the \textit{else} block of the \textit{try/except} statement. This ensures the update only occurs when no exception was raised.
\item \textbf{Patch 3} (correct): Also uses a temporary variable, but performs the assignment inside the \textit{try} block immediately. Since execution halts immediately if an exception is raised, this also prevents the previous bug.
\end{itemize}

In this case, R4P correctly identifies Patches 2 and 3 as correct patch, yet incorrectly consider Patch 4 as well. In contrast, the two Claude models only correctly identify either Patch 2 or 3.

Below are the raw content of the issue and patches, as well as the reasoning trajectory and corresponding analysis of R4P, Claude-3.7-Sonnet, and Claude-4.0-Sonnet.

\begin{lstlisting}
<issue>
MiddlewareNotUsed leaves undesired side effects when loading middleware in ASGI context
Description

I experienced strange issues when working with ASGI, django-debug-toolbar and my own small middleware. It was hard problem to debug, I uploaded an example project here: https://github.com/hbielenia/asgi-djangotoolbar-bug (the name is misleading - I initially thought it's a bug with django-debug-toolbar).
The SESSION_FILE_PATH setting is intentionally broken to cause a 500 error. When starting the application and accessing /admin (any location really, but I wanted to leave it at a minimum and didn't add any views) it gives TypeError: object HttpResponse can't be used in 'await' expression. Commenting out asgi_djangotoolbar_bug.middleware.DummyMiddleware fixes the issue (in that I receive a 500 ImproperlyConfigured exception). I'm not sure about the overall role of django-debug-toolbar here - removing it causes Daphne to return a 500 error page but without debug information and there's no traceback in console either. I decided to leave it since it helped me approximate the causes of issue.
I notice that in https://github.com/django/django/blob/3.1.4/django/core/handlers/base.py#L58 while MiddlewareNotUsed causes the loop to skip futher processing and go to next middleware, it does leave handler variable overwritten with output of self.adapt_method_mode(). On next pass, this handler is passed to next middleware instance, disregarding all the previous checks for (lack of) async support. This likely causes the middleware chain to be "poisoned" from this point onwards, resulting in last middleware in response cycle to return an HttpResponse as a synchronous middleware would, instead of coroutine that is expected.
This is probably avoided by adding async support to my middleware, but unless I'm missing something docs indicate it should work as it is. It is my intention that it's applied only on synchronous requests, so I didn't make it async compatible on purpose. If it's intentional in Django that every middleware needs to support async if the application is run as ASGI app, the documentation should probably state that clearly. Though it kinda defeats the purpose of having async_capable = False flag in the first place.
</issue>

<patch-1>
diff --git a/django/core/handlers/base.py b/django/core/handlers/base.py
--- a/django/core/handlers/base.py
+++ b/django/core/handlers/base.py
@@ -51,11 +51,12 @@ class BaseHandler:
                 middleware_is_async = middleware_can_async
             try:
                 # Adapt handler, if needed.
+                # Adapt handler, if needed.
+                mw_instance = middleware(handler)
                 handler = self.adapt_method_mode(
                     middleware_is_async, handler, handler_is_async,
                     debug=settings.DEBUG, name='middleware %s' % middleware_path,
                 )
-                mw_instance = middleware(handler)
             except MiddlewareNotUsed as exc:
                 if settings.DEBUG:
                     if str(exc):
</patch-1>
<patch-2>
diff --git a/django/core/handlers/base.py b/django/core/handlers/base.py
--- a/django/core/handlers/base.py
+++ b/django/core/handlers/base.py
@@ -51,11 +51,11 @@ class BaseHandler:
                 middleware_is_async = middleware_can_async
             try:
                 # Adapt handler, if needed.
-                handler = self.adapt_method_mode(
+                adapted_handler = self.adapt_method_mode(
                     middleware_is_async, handler, handler_is_async,
                     debug=settings.DEBUG, name='middleware %s' % middleware_path,
                 )
-                mw_instance = middleware(handler)
+                mw_instance = middleware(adapted_handler)
             except MiddlewareNotUsed as exc:
                 if settings.DEBUG:
                     if str(exc):
@@ -63,6 +63,9 @@ class BaseHandler:
                     else:
                         logger.debug('MiddlewareNotUsed: %r', middleware_path)
                 continue
+            else:
+                # Only update handler if middleware was not skipped
+                handler = adapted_handler
 
             if mw_instance is None:
                 raise ImproperlyConfigured(
</patch-2>

<patch-3>
diff --git a/django/core/handlers/base.py b/django/core/handlers/base.py
--- a/django/core/handlers/base.py
+++ b/django/core/handlers/base.py
@@ -50,12 +50,14 @@ class BaseHandler:
             else:
                 middleware_is_async = middleware_can_async
             try:
-                # Adapt handler, if needed.
-                handler = self.adapt_method_mode(
+                # Only adapt handler if the middleware is used
+                adapted_handler = self.adapt_method_mode(
                     middleware_is_async, handler, handler_is_async,
                     debug=settings.DEBUG, name='middleware %s' % middleware_path,
                 )
-                mw_instance = middleware(handler)
+                mw_instance = middleware(adapted_handler)
+                # If we get here, the middleware will be used, so update the handler
+                handler = adapted_handler
             except MiddlewareNotUsed as exc:
                 if settings.DEBUG:
                     if str(exc):
</patch-3>

<patch-4>
diff --git a/django/core/handlers/base.py b/django/core/handlers/base.py
--- a/django/core/handlers/base.py
+++ b/django/core/handlers/base.py
@@ -50,11 +50,7 @@ class BaseHandler:
             else:
                 middleware_is_async = middleware_can_async
             try:
-                # Adapt handler, if needed.
-                handler = self.adapt_method_mode(
-                    middleware_is_async, handler, handler_is_async,
-                    debug=settings.DEBUG, name='middleware %s' % middleware_path,
-                )
+                # Create middleware instance first to check if it will be used
                 mw_instance = middleware(handler)
             except MiddlewareNotUsed as exc:
                 if settings.DEBUG:
@@ -64,6 +60,12 @@ class BaseHandler:
                         logger.debug('MiddlewareNotUsed: %r', middleware_path)
                 continue
 
+            # Only adapt the handler if the middleware will be used
+            handler = self.adapt_method_mode(
+                middleware_is_async, handler, handler_is_async,
+                debug=settings.DEBUG, name='middleware %s' % middleware_path,
+            )
+
             if mw_instance is None:
                 raise ImproperlyConfigured(
                     'Middleware factory %s returned None.' % middleware_path
</patch-4>
\end{lstlisting}

\paragraph{Reasoning of R4P}

R4P correctly finds that Patch 2 and Patch 3 achieve the fix with a temporary variable and conditionally updating the main handler. However, R4P hallucinates regarding Patch 4. It considers Patch 4 ensures the handler is correctly adapted before used by the middleware instance, but actually this handler is unadapted.

\begin{lstlisting}
Let's analyze each patch to determine which one(s) can resolve the issue in the `MiddlewareNotUsed` handling in the ASGI context.

### Issue Analysis
The issue is that when `MiddlewareNotUsed` is raised, the `handler` variable is being overwritten with the output of `self.adapt_method_mode()`, which can cause issues in the middleware chain. Specifically, it can lead to a `TypeError` when the last middleware in the chain returns an `HttpResponse` instead of a coroutine.

### Patch Analysis

#### Patch-1
- Adds a comment to explain the adaptation of the handler.
- Moves the `mw_instance = middleware(handler)` line after the adaptation of the handler.
- This patch ensures that the `handler` is adapted before being passed to the middleware instance.
- This is a good change because it ensures that the `handler` is correctly adapted before being used by the middleware instance.

#### Patch-2
- Adds a comment to explain the adaptation of the handler.
- Moves the `mw_instance = middleware(handler)` line after the adaptation of the handler.
- Adds an `else` block to update the `handler` only if the middleware is not skipped.
- This patch ensures that the `handler` is correctly adapted before being passed to the middleware instance and only updates the `handler` if the middleware is not skipped.
- This is a good change because it ensures that the `handler` is correctly adapted before being used by the middleware instance and prevents the `handler` from being updated if the middleware is skipped.

#### Patch-3
- Adds a comment to explain the adaptation of the handler.
- Adds a comment to explain that the handler is only adapted if the middleware is used.
- Moves the `mw_instance = middleware(handler)` line after the adaptation of the handler.
- Adds a comment to explain that the handler is updated only if the middleware is used.
- This patch ensures that the `handler` is correctly adapted before being passed to the middleware instance and only updates the `handler` if the middleware is not skipped.
- This is a good change because it ensures that the `handler` is correctly adapted before being used by the middleware instance and prevents the `handler` from being updated if the middleware is skipped.

#### Patch-4
- Moves the `mw_instance = middleware(handler)` line before the adaptation of the handler.
- Adds a comment to explain that the middleware instance is created first to check if it will be used.
- Adds a comment to explain that the handler is only adapted if the middleware will be used.
- This patch ensures that the `handler` is correctly adapted before being passed to the middleware instance and only updates the `handler` if the middleware is not skipped.
- This is a good change because it ensures that the `handler` is correctly adapted before being used by the middleware instance and prevents the `handler` from being updated if the middleware is skipped.

### Conclusion
Patches 2, 3, and 4 are all good changes. They ensure that the `handler` is correctly adapted before being passed to the middleware instance and only updates the `handler` if the middleware is not skipped. Patch-1 is also a good change, but it doesn't add the `else` block to update the `handler` only if the middleware is not skipped.

Therefore, the correct patches are:

\boxed{2, 3, 4}
\end{lstlisting}

\paragraph{Reasoning of Claude-3.7-Sonnet}

Claude-3.7 correctly identifies Patch but rejects Patch 3 due to a fundamental misunderstanding of exception control flow. It states that Patch 3 ``updates handler before the except block, which means if \textit{MiddlewareNotUsed} is raised, the change to handler will be lost." However, if the line \textit{mw\_instance = middleware(...)} raises an exception, the immediately following line \textit{handler = adapted\_handler} is never executed. Thus, Patch 3 functionally equivalent to Patch 2. The model hallucinated a side effect where the assignment occurs despite the exception interrupting the flow.

\begin{lstlisting}
I need to analyze the issue and the proposed patches to determine which one(s) correctly fix the problem.

## Understanding the Issue
The issue occurs when `MiddlewareNotUsed` is raised in an ASGI context. Currently, the code adapts the handler first (via `self.adapt_method_mode()`), and then creates the middleware instance. If `MiddlewareNotUsed` is raised when creating the middleware instance, the adapted handler still remains and is passed to the next middleware, which causes type mismatches when mixing sync and async middleware.

The key problem is that `handler` is modified before we know if the middleware will be used or not. If the middleware isn't used (raises `MiddlewareNotUsed`), we've already changed `handler` to be sync or async based on the middleware that ended up not being used.

## Patch Analysis

### Patch 1
```python
# Adapt handler, if needed.
# Adapt handler, if needed. (duplicate comment)
mw_instance = middleware(handler)
handler = self.adapt_method_mode(
    middleware_is_async, handler, handler_is_async,
    debug=settings.DEBUG, name='middleware %s' % middleware_path,
)
```

This patch switches the order of operations - it creates the middleware instance first, and only then adapts the handler. This approach would fail because the middleware constructor expects to receive a handler that's compatible with its sync/async nature, but we're passing the potentially incompatible handler directly.

### Patch 2
```python
adapted_handler = self.adapt_method_mode(
    middleware_is_async, handler, handler_is_async,
    debug=settings.DEBUG, name='middleware %s' % middleware_path,
)
mw_instance = middleware(adapted_handler)
...
else:
    # Only update handler if middleware was not skipped
    handler = adapted_handler
```

This patch computes the adapted handler but doesn't update the main `handler` variable until we're sure the middleware isn't skipped. It passes the adapted handler to the middleware instance, and only if no `MiddlewareNotUsed` is raised does it update `handler` to be the adapted version. This correctly addresses the issue.

### Patch 3
```python
adapted_handler = self.adapt_method_mode(
    middleware_is_async, handler, handler_is_async,
    debug=settings.DEBUG, name='middleware %s' % middleware_path,
)
mw_instance = middleware(adapted_handler)
# If we get here, the middleware will be used, so update the handler
handler = adapted_handler
```

Similar to Patch 2, this uses a temporary variable for the adapted handler, but it updates `handler` before the `except` block, which means if `MiddlewareNotUsed` is raised, the change to `handler` will be lost. This doesn't fix the issue.

### Patch 4
```python
# Create middleware instance first to check if it will be used
mw_instance = middleware(handler)
...
# Only adapt the handler if the middleware will be used
handler = self.adapt_method_mode(
    middleware_is_async, handler, handler_is_async,
    debug=settings.DEBUG, name='middleware %s' % middleware_path,
)
```

This patch tries to create the middleware instance first using the unadapted handler, then only adapts the handler if the middleware will be used. However, as noted for Patch 1, this would likely fail because the middleware constructor needs a compatible handler.

## Conclusion

After analyzing all patches, I determine that only Patch 2 correctly fixes the issue by:
1. Creating a temporary adapted handler
2. Using it to initialize the middleware
3. Only updating the main handler variable if the middleware isn't skipped

The other patches either introduce new issues or don't fully address the original problem.

\boxed{2}
\end{lstlisting}
\paragraph{Reasoning of Claude-4.0-Sonnet}

Claude-4.0 correctly analyzes the logic of Patch 2 and 3 (but also incorrectly consider Patch 4 as correct at its initial analysis), and even explicitly acknowledges that Patch 2 ``addresses the core issue" and ``should work." Then, in the final comparison of Patch 2,3,4, it correctly filters out Patch 4. However, it also dismisses Patch 2 merely because the \textit{else} block adds ``extra nesting" and Patch 3 appeared ``cleaner." Consequently, it generated a false negative for Patch 2. Interestingly, it is aware of the final objective is to find all patch(es) at the very begining of the reasoning, but soon forgot this goal after a very long thought, and tend to find the ``best" patch. We consider it is a problem caused by long-term forgetting.

\begin{lstlisting}
<think>
Let me understand the issue first:

1. The problem occurs when using ASGI with middleware that raises `MiddlewareNotUsed` exception
2. In the current code, when `MiddlewareNotUsed` is raised, the `handler` variable has already been modified by `self.adapt_method_mode()` but the middleware is skipped
3. This "poisoned" handler is then passed to the next middleware, causing async/sync mismatch issues
4. The result is that synchronous middleware returns `HttpResponse` instead of the expected coroutine

Let me examine the current code flow:
```python
try:
    # Adapt handler, if needed.
    handler = self.adapt_method_mode(...)  # This modifies handler
    mw_instance = middleware(handler)      # This might raise MiddlewareNotUsed
except MiddlewareNotUsed as exc:
    # handler is already modified at this point!
    continue
```

The problem is that `handler` gets modified even when the middleware will be skipped due to `MiddlewareNotUsed`.

Now let me analyze each patch:

**Patch 1:**
```python
# Adapt handler, if needed.
# Adapt handler, if needed.  # Duplicate comment
mw_instance = middleware(handler)
handler = self.adapt_method_mode(...)
```
This moves the middleware instantiation before the handler adaptation, but it doesn't solve the core issue. If `MiddlewareNotUsed` is raised during middleware instantiation, the handler won't be adapted at all, which could break the chain. This doesn't look right.

**Patch 2:**
```python
adapted_handler = self.adapt_method_mode(...)
mw_instance = middleware(adapted_handler)
# ... exception handling ...
else:
    # Only update handler if middleware was not skipped
    handler = adapted_handler
```
This creates a temporary `adapted_handler`, uses it to create the middleware instance, and only assigns it back to `handler` if no exception was raised (in the `else` block of the try/except). This looks good and addresses the core issue.

**Patch 3:**
```python
adapted_handler = self.adapt_method_mode(...)
mw_instance = middleware(adapted_handler)
# If we get here, the middleware will be used, so update the handler
handler = adapted_handler
```
This is similar to Patch 2, but updates the handler immediately after successful middleware instantiation, still within the try block. This also addresses the issue but is slightly different in timing.

**Patch 4:**
```python
try:
    # Create middleware instance first to check if it will be used
    mw_instance = middleware(handler)
except MiddlewareNotUsed as exc:
    # ... handle exception ...
    continue

# Only adapt the handler if the middleware will be used
handler = self.adapt_method_mode(...)
```
This completely reverses the order - first creates the middleware instance with the original handler, and only adapts the handler if the middleware will be used. This could be problematic because the middleware might expect an adapted handler during instantiation.

Let me think about which approach is best:

1. **Patch 1** is clearly wrong - it moves the adaptation after middleware creation, which could break things.

2. **Patch 4** might have issues because middleware instantiation happens with the non-adapted handler, but then the adapted handler is used later. This could cause inconsistencies.

3. **Patches 2 and 3** both use the correct approach: adapt the handler first (to a temporary variable), try to create the middleware with the adapted handler, and only commit the adaptation if successful.

Between Patch 2 and Patch 3:
- Patch 2 uses the `else` clause of try/except, which is more explicit about "only if no exception occurred"
- Patch 3 updates the handler immediately after successful middleware creation, still in the try block

Both should work, but Patch 2 is slightly more explicit and follows the pattern of "only do this if the try block succeeded completely". The `else` clause in try/except is specifically designed for this use case.

However, let me double-check the logic. In both patches 2 and 3, if `MiddlewareNotUsed` is raised during `middleware(adapted_handler)`, the original `handler` remains unchanged, which is exactly what we want.

Patch 3 is actually more straightforward and easier to read - if we successfully create the middleware instance, we immediately update the handler. The else clause in Patch 2 adds an extra level of nesting that might not be necessary.

Actually, let me reconsider Patch 4. The issue might be more subtle. Let me think about when `MiddlewareNotUsed` is typically raised:

`MiddlewareNotUsed` is usually raised in the middleware's `__init__` method when it determines it shouldn't be used. At this point, the middleware examines the handler that's passed to it. If the handler adaptation is important for the middleware to make its decision, then Patch 4 could be problematic.

However, if the middleware's decision to raise `MiddlewareNotUsed` is independent of the handler's adaptation (which is likely the case in most scenarios), then Patch 4 could actually work.

But looking at the specific issue described, it seems like the problem is that the handler adaptation happens regardless of whether the middleware will be used, so the safer approach is Patches 2 or 3.

Between Patch 2 and 3, I think Patch 3 is cleaner and more direct. The comment also makes the intent very clear.
</think>

Looking at this issue, I need to understand the core problem: when `MiddlewareNotUsed` is raised, the `handler` variable has already been modified by `self.adapt_method_mode()`, but the middleware is skipped. This "poisoned" handler then gets passed to subsequent middleware, causing async/sync compatibility issues.

Let me analyze each patch:

**Patch 1**: This moves middleware instantiation before handler adaptation, but has issues:
- The duplicate comment suggests it's incomplete
- More importantly, it reverses the logical order - middleware typically expects an adapted handler
- If `MiddlewareNotUsed` is raised, the handler won't be adapted at all, which could break the middleware chain

**Patch 2**: This uses a temporary variable approach:
- Creates `adapted_handler` first
- Uses it for middleware instantiation  
- Only updates the main `handler` in the `else` clause if no exception occurred
- This properly isolates the adaptation from the exception handling

**Patch 3**: Similar to Patch 2 but simpler:
- Creates `adapted_handler` and uses it for middleware instantiation
- Updates `handler` immediately after successful middleware creation
- The comment clearly explains the intent
- More direct and easier to follow than Patch 2

**Patch 4**: This completely reverses the order:
- Creates middleware instance first with the original handler
- Only adapts the handler if middleware will be used
- This could be problematic because middleware might need the adapted handler during instantiation to make decisions

Patches 2 and 3 both solve the core issue correctly by preventing handler modification when middleware is skipped. Between them, Patch 3 is cleaner and more straightforward - it updates the handler immediately after successful middleware creation, with a clear comment explaining the logic.

Patch 4 risks breaking middleware that depends on receiving an adapted handler during instantiation, and Patch 1 has the wrong order and incomplete implementation.
\boxed{3}
\end{lstlisting}

\end{document}